\documentclass[review]{elsarticle}

\usepackage{natbib}
\usepackage{graphicx}
\usepackage{amsmath}
\graphicspath{{Images/}}

\begin{document}

\begin{frontmatter}

\title{Simple Two-Dimensional Object Tracking based on a Graph Algorithm}

%% Group authors per affiliation:
\author{Alexandra Heidsieck}
\address{Zentralinstitut f\"ur Medizintechnik\\
           Technische Universit\"at M\"unchen}
\cortext[mycorrespondingauthor]{Corresponding author}
\ead{aheidsieck@tum.de}

\begin{abstract}
The visual observation and tracking of cells and other micrometer-sized objects has many different biomedical applications. The automation of those tasks based on computer methods helps in the evaluation of such measurements. In this work, we present a general purpose algorithm that excels at evaluating deterministic behavior of micrometer-sized objects. Our concrete application is the tracking of fast moving objects over large distances along deterministic trajectories in a microscopic video. Thereby, we are able to determine characteristic properties of the objects. For this purpose, we use a set of basic algorithms, including blob recognition, feature-based shape recognition and a graph algorithm, and combined them in a novel way. An evaluation of the algorithms performance shows a high accuracy in the recognition of objects as well as of complete trajectories. Moreover, a direct comparison to a similar algorithm shows superior recognition rates.
\end{abstract}


\end{frontmatter}

\section{Introduction}\label{sec:intro}

Microscopic cell imaging is a standard technique to gain information about the life cycle and movement of cells and objects of similar size. However, the observation of the temporal or spatial development of those objects over a long measurement period can be a difficult and time consuming task if carried out manually. The automated observation and tracking based on computer methods helps in the evaluation of such measurements, e.g. the long time observation of cells for the purpose of activity, proliferation, apoptosis and/or translocation. Apart from the study of living cells and their biological behaviors, there are other objects and properties which can also be of interest for various applications. One property is the characteristic object velocity in the presence of a certain type of force field, e.g. an electric or magnetic field. Here, the focus of the automated tracking algorithm lies on slightly different properties compared to the observation of biological processes.\\

A cell or particle tracking algorithm usually consists of two main parts. The first part is the identification and recognition of the objects and the association of certain properties with the objects (segmentation). The second part is the association of those objects with each other over a set of frames and the reconstruction of the actual trajectory (linking).\\
Often used approaches for the recognition and identification of objects, are template matching \citep{brunelli} or watershed transformation \citep{Vincent:1991, Roerdink:2000}.\\
The simplest approach to connect objects over two fra\-mes is to associate each object with the closest object in the next frame. Here, "closest" can refer to the spatial distance as well as to object similarities. A purely spatial connection of recognized objects may work well for sparsely populated frames, but fails for higher object densities. Depending on the final application, the consideration of object features such as shape, size or luminosity may be useful or disadvantageous.\\

Though there are many particle and cell tracking algorithms freely or commercially available \citep[and references therein]{Meijering:2012}, most of those \citep[e.g.][]{Jaqaman:2008,Huth:2011} focus on cell specific behavior, e.g. tracking Brownian motion or morphodynamic behavior \citep[e.g.][]{Mosig:2009,Huth:2011}. They are often meant to track only small distances and non-overlapping trajectories and, therefore, often perform poorly when applied to problems with large distances between individual particle locations. Other packages are able to track in three dimensions \citep[e.g.][]{Leocmach:2013,Dzyubachyk:2010} or are only available together with a physical measurement set-up.\\
Therefore, we created our own cell and microbubble tracking algorithm which is able to reliably track fast moving objects and includes a feature based recognition of those objects. In the following, we introduce a general purpose algorithm that excels at evaluating deterministic behavior of micrometer-sized objects. Our concrete application is the tracking of fast moving objects over large distances along deterministic trajectories. However, changing certain key aspects of our evaluation, e.g. the cost or weight functions for the graph algorithm, offers the possibility to adapt the algorithm to almost any tracking problem. We focus our discussion on the problem at hand but point out possible modifications for different applications.\\

The investigated cells or microbubbles are loaded with magnetic nanoparticles as described e.g. by \citeauthor{Kilgus:2012} \citep{Kilgus:2012} or \citeauthor{Mannell:2012} \citep{Mannell:2012} and placed inside a well-defined magnetic field. The objects consequently move in approximately the same direction towards the magnetic field source. A sequence of microscopic images of this movement is recorded and subsequently evaluated with the presented algorithm. Based on the object size and measured velocity, certain characteristic properties of the objects like magnetophoretic mobility and magnetic moment \citep{zborowski,Chalmers:1999,Haefeli:2005} can be inferred.

\section{Method}

The presented algorithm consists of several steps which can be grouped roughly into four parts. In the first part, the image data is loaded and filtered. It is based on the IDL particle tracking algorithm by \citeauthor{Crocker:1996} \citep{Crocker:1996, Crocker:1996:online} and its Matlab adaption by \citeauthor{Blair:online} \citep{Blair:online} and is explained in step 1. In the second part, all object positions in all images are found. This is represented by steps 2 and 3. In the next part, the connections between the individual objects in the images are found and the trajectories are reconstructed in steps 4 to 6. Finally, in the last part (step 7), the trajectory data is evaluated. The individual steps of the algorithm are explained in more detail in the following.

\begin{figure*}
	\includegraphics[width=\textwidth]{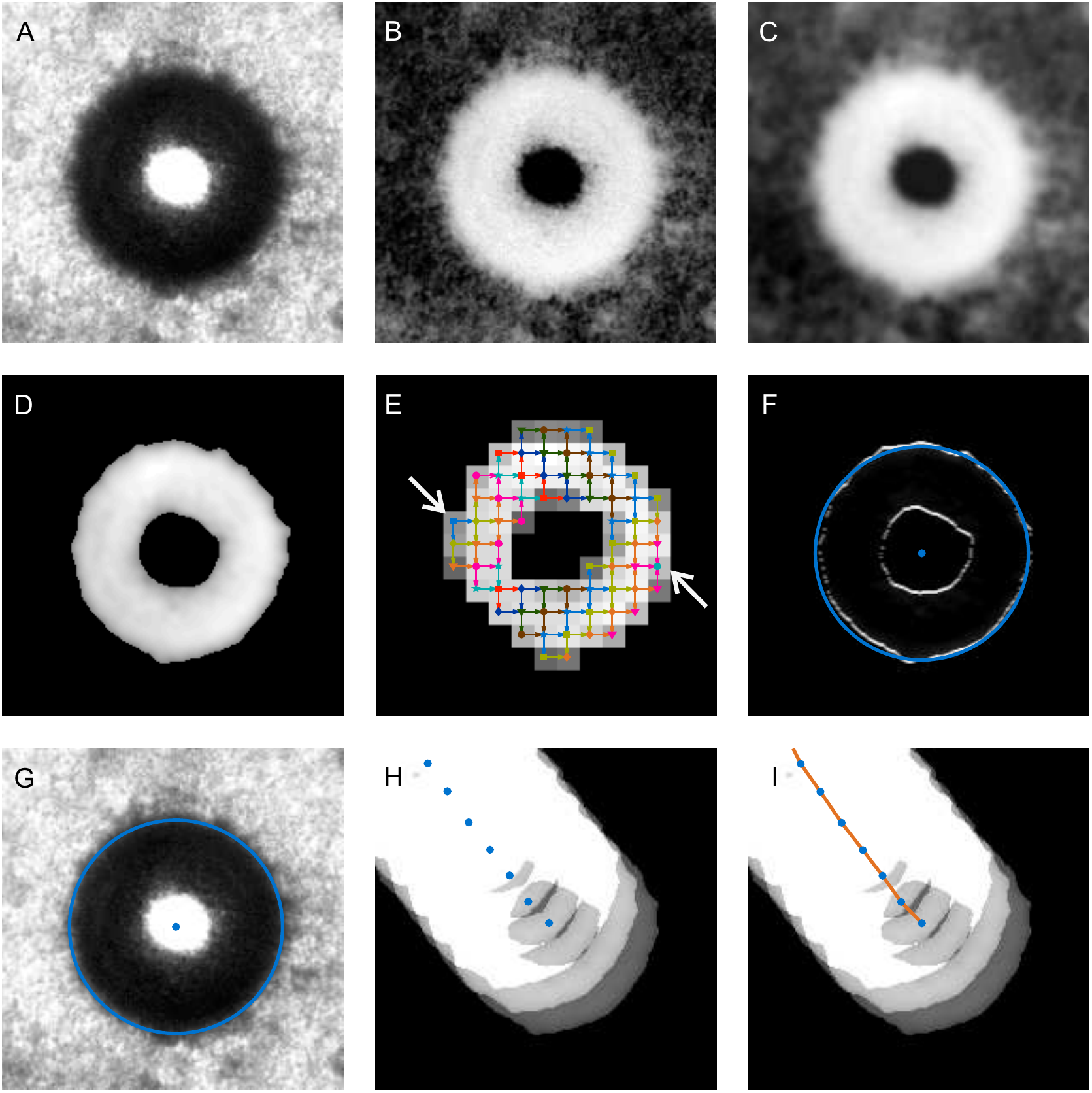}
	\caption[Individual steps of the tracking algorithm on the example of a microbubble]{Individual steps of the tracking algorithm on the example of a microbubble; the frame is imported (A), if necessary inverted (B), filtered (C) and a threshold is applied (D); subsequently the object positions are located (E) with centroid and radius (F,\,G) collected over all frames (H) and the trajectories are reconstructed (I)}
	\label{fig:algo_steps}
\end{figure*}%
\begin{enumerate}
	\item The selected frames of the image sequence are imported and, if necessary, converted into gray scale. An example frame is shown in figure \ref{fig:algo_steps}A. Depending on the coloring of the frame, the color spectrum has to be inverted to achieve bright objects on a dark background, e.g. in case of microbubble microscopy (figure \ref{fig:algo_steps}B). Based on the approximate object size $W$, e.g. obtained by the microscopy software, and a freely chosen noise level $N$, the frames are band-pass filtered. The filter applies convolutions of the frame matrix with a gaussian and a rectangular filter
	\begin{align}
		F_\mathrm{gauss}(\vec{x})	&= \frac{1}{f_\mathrm{gauss}} \exp{-\frac{\vec{x}^2}{2N^2}}\,\label{equ:kernel}\\
		F_\mathrm{rect}(\vec{x})		&= \frac{1}{f_\mathrm{rect}} \text{rect}\left(2W + 1\right)\notag\\
								&= \left\{ \begin{array}{cl}\frac{1}{2W + 1} & \text{if } -\frac{2W+1}{2} \leq | \vec{x} | \leq \frac{2W+1}{2}\\
													0 & \text{else}
							\end{array}\right.\,
	\end{align}
	where $f_\mathrm{gauss, rect}$ are normalization factors and \emph{rect} is the rectangular or boxcar function. The difference between applying both filters individually smoothes the frame and subtracts the background \citep{Crocker:1996, Crocker:1996:online, Blair:online}. The resulting frame is shown in figure \ref{fig:algo_steps}C. Afterwards, we apply a freely chosen threshold, see figure \ref{fig:algo_steps}D.

	\item In the next step, the objects are located with the help of a connected-component algorithm \citep{Davies, Cormen, Parker}. Here, all connected pixels are collected and assigned to an object, see figure \ref{fig:algo_steps}E. To better illustrate the process, the example frame was rescaled to represent an object with only a few pixels in width.\\
	To locate the connected pixels, the frame is scanned for non-zero pixels. If such a pixel is found, it is marked with an identification number and subsequently removed from the frame. This is repeated for the four cartesian neighbors of the original pixel until no neighbors with a non-zero value remain. In figure \ref{fig:algo_steps}E, the start pixel in the upper left corner is indicated by a white arrow. Subsequently, all neighboring pixels are found, as indicated by the different symbols. The colored arrows represent the individual steps where the non-zero neighbors are located until the final pixel is found, again indicated by a white arrow. All pixels marked with a certain identification number then belong to the same connected object. The next found object receives a new identification number. Overlapping and touching objects are recognized as one larger object as long as they are directly connected by at least one pixel.\\% In contrast to the often described version on binary data, we perform this algorithm on the thresholded gray scale datasets. This offers further possibilities in distinguishing touching objects.\\
	In the following, $P_n$ denotes the set of all pixels $k$ associated with the object with the identification number $n$, while $B_m$ denotes the individual frames. The intensity-weighted centroid $\vec{c}_n$ of the found objects $n$ and the euclidian distance $r_k$ from each pixel $k$ of the object $P_n$ to the centroid
	\begin{align}
		\vec{c}_n &= \frac{\sum\limits_{k} I(\vec{x}_k) \vec{x}_k}{\sum\limits_{k} I(\vec{x}_k)}\,,\\
		r_k &= \| \vec{c}_n - \vec{x}_k \|\,,
	\end{align}
	are calculated for all $k \in P_n$. Here, $I(\vec{x}_k)$ is the intensity of the pixel $k$ at the position described by $\vec{x}_k$.
	To calculate the object radius, a sobel filter is applied. Thereby, the edge of the object is obtained as shown in figure \ref{fig:algo_steps}F. On the sobel-filtered frame $S_m$, we calculate the radius $R_n$ of the object
	\begin{align}
		R_n = \frac{\sum\limits_k I_\mathrm S(r_k) r_k}{\sum\limits_k I_S(r_k)}\,,
	\end{align}
	by calculating the weighted arithmetic mean of the distance $r_{k}$ of the (outer) edge from the centroid for \mbox{$k \in \left\{k \in S_m |\, \overline{r_k} < r_k < \max{r_k} \right\}$}. The calculation of the mean value is again weighted by the intensity, or rather the intensity $I_S$ of the sobel filtered frame $S_m$ which can be interpreted as the intensity gradient of the original frame. We limit the selection of edge pixels, on the one hand, by taking only those pixels into account which have a larger distance than the mean distance of all object pixels. Thereby, we discard the inner edge in case of microbubbles. On the other hand, we remove those edge pixels at a larger distance than the maximum distance of the object pixels, thereby limiting the influence of other objects in close proximity. The resulting centroid and radius are also shown in figure \ref{fig:algo_steps}F in reference to the sobel filtered image and in figure \ref{fig:algo_steps}G in reference to the original microscope image.

	\item Finally, we filter the found objects according to their size and shape. Up to date, we included two shapes. The first shape is simply a circular object which we applied to images of cells as shown in figure \ref{fig:microbubble_tracking}A, the second shape is a circular ring shape which was applied to microbubbles, compare figure \ref{fig:microbubble_tracking}B.\\
	The found objects are on the one hand filtered by allowing only a certain size range% and a certain variation $\Delta_r$ of the pixel distance from the centroid
	\begin{align}
		%~ \frac{r_k - \overline{r}_i}{\sqrt{\sum\limits_k(r_k - \overline{r}_i)^2}} > \Delta_r\,,&& \text{where } k \in P_i \label{equ:blobfilter_a}\\
		R_\mathrm{min} \leq R_n \leq R_\mathrm{max}\,. \label{equ:blobfilter_b}
	\end{align}
	On the other hand, we also allow only those objects which fit our shape-based features. In case of circles, the total area of those objects has to be in good agreement with their reconstructed radius (circularity condition)
	\begin{align}
		\left| \frac{R_n^2\pi}{N_k} - 1 \right|	&< \Delta_A\,,
	\end{align}
	where $N_k$ is the number of pixels $k \in P_n$. In case of rings, we limit the objects to those with a black center of minimum size
	\begin{align}
		\frac{r_k}{\mathrm{min}\left(k\right)}					&< \Delta_C\,.
	\end{align}
	Additionally, we allow only those objects with a high intensity weighted normalized match index \citep{Parker}
	\begin{align}
		\frac{\overline{I}(k_\text{in}) - \overline{I}(k_\text{out}) - N_{0,\text{in}}\cdot\overline{I}(k)}{\overline{I}(k_\text{in}) + \overline{I}(k_\text{out}) + N_{0,\text{in}}\cdot\overline{I}(k)}	&> \Delta_I\,,
		%\frac{\overline{I}(k_{R})}{\overline{I}(k)}	&> \Delta_I\,,
	\end{align}
	where $I(k_\text{in})$ describes the intensity of the pixels $k_\text{in} \in \left\{k \in P_n |\, r_k < R_n\right\}$ in case of circles or $k_\text{in}\in \left\{k \in P_n |\, R_n/4 < r_k < R_n\right\}$ in case of rings inside a circle with the radius $R_n$. $k_\text{out}$ describes the pixels outside of a circle with the radius $R_n$ in a similar manner. $N_{0,\text{in}}$ denotes the number of empty pixels $k_\text{in}$ inside the radius. Thereby, we ensure a good agreement of our objects with circles or rings.\\
	Figure \ref{fig:microbubble_tracking}C exemplarily shows a microscopy image and the objects found by our algorithm with their corresponding size denoted as blue circles. Typical values for the maximum/minimum allowed parameters $\Delta$ can be found in table \ref{tab:cell_params2}.
	
	\item After the object reconstruction, the detected objects are then collected into a single dataset including a reference to the originating frame. A set of several identified objects in close proximity is shown as blue dots in figure \ref{fig:algo_steps}H, including an overlay of the corresponding filtered frames. For the purpose of the reconstruction of the trajectory, we beforehand calculate the costs for the connections between all objects located in neighboring frames $B_m$ and $B_{m+1}$. Our cost functions for those edges include the absolute distance $s_{np}$ and the relative radius $\Delta R_{np}$ of any two objects $n$ and $p$
	\begin{align}
		s_{np}			&=	\| \vec{x}_n - \vec{x}_p \|\,,		\label{equ:weights1}\\
		\Delta R_{np}	&=	\left| \frac{R_n}{R_p} - 1\right|\,,	\label{equ:weights2}
	\end{align}
	where $n \in B_{m}$ and $p \in B_{m+1}$. To judge the angle of the trajectory, we first calculate the angle between all objects of neighboring frames relative to the cartesian coordinates $x$ and $y$. The relative angle $\Delta\varphi_{np}$ for each edge is then defined in reference to the most often occurring angle $\Phi$ over the complete frame set
	\begin{align}
		\varphi_{np}		&=	\mathrm{atan}\left(\frac{y_n-y_p}{x_n-x_p}\right)\,,	\label{equ:weights3}\\
		\Phi				&=	\mathrm{max}\left(H_{s_{np}}\left(\varphi_{np}\right)\right) \text{mod\ } 180^\circ\,,		\label{equ:weights4}\\
		\Delta\varphi_{np}	&=	\frac{\varphi_{np}}{\Phi}\,,		\label{equ:weights5}
	\end{align}
	where $n \in B_{m}$ and $p \in B_{m+1}$ and $H_{s_{np}}(\varphi_{np})$ denotes the histogram over the angles $\varphi_{np}$ weighted with the distances $s_{np}$. However, we cannot distinguish between backward and forward angles, since the direction of the previous edge is not yet known.
%\begin{figure}[t]
	%\centering
	%\includegraphics[scale=.5]{graph_algo}
	%\caption{Schematic description of graph algorithm for a sequence of three subsequent frames (a-c); arrows denote edges between objects with their according weights noted above them; panel (d) shows an overlay of panel (a) to (c)}
	%\label{fig:graph_algo}
%\end{figure}%

\begin{table}
	\begin{center}
	\begin{tabular}{cc}
		\hline
		Parameter	& Value\\	\hline
		%~ $\Delta_r$	&	0.1\\
		$\Delta_C$	&	5\\
		$\Delta_A$	&	0.3\\
		$\Delta_I$	&	0.5\\	\hline
	\end{tabular}
	\caption{Typical parameters used for peak identification (compare step 3) of trajectory reconstruction}
	\label{tab:cell_params2}
	\end{center}
\end{table}%
\begin{table}
	\begin{center}
	\begin{tabular}{lc}
		\hline
		Parameter & Value\\ \hline
		Minimum diameter		& $0.5\cdot w$\\
		Maximum distance		& $10\cdot w$\\
		Minimum track length	& $5$\\
		Graph weight distance $G_s$		& $1$\\
		Graph weight radius $G_r$		& $1$\\
		Graph weight angle $G_\varphi$	& $2$\\ \hline
	\end{tabular}
	\caption{Typical parameters for user input}
	\label{tab:cell_params1}
	\end{center}
\end{table}%
\begin{table}
	\begin{center}
	\begin{tabular}{lc}
		\hline
		Parameter & Value \\	\hline
		Maximum angle			& $30^\circ$\\
		Maximum angle std		& $45^\circ$\\
		Maximum radius std		& $0.5$\\
		Maximum distance std	& $0.5$\\	\hline
	\end{tabular}
	\caption{Typical parameters used for optimization (compare step 6) of trajectory reconstruction}
	\label{tab:cell_params3}
	\end{center}
\end{table}

\begin{figure}
	\centering
	\includegraphics[width=.5\textwidth]{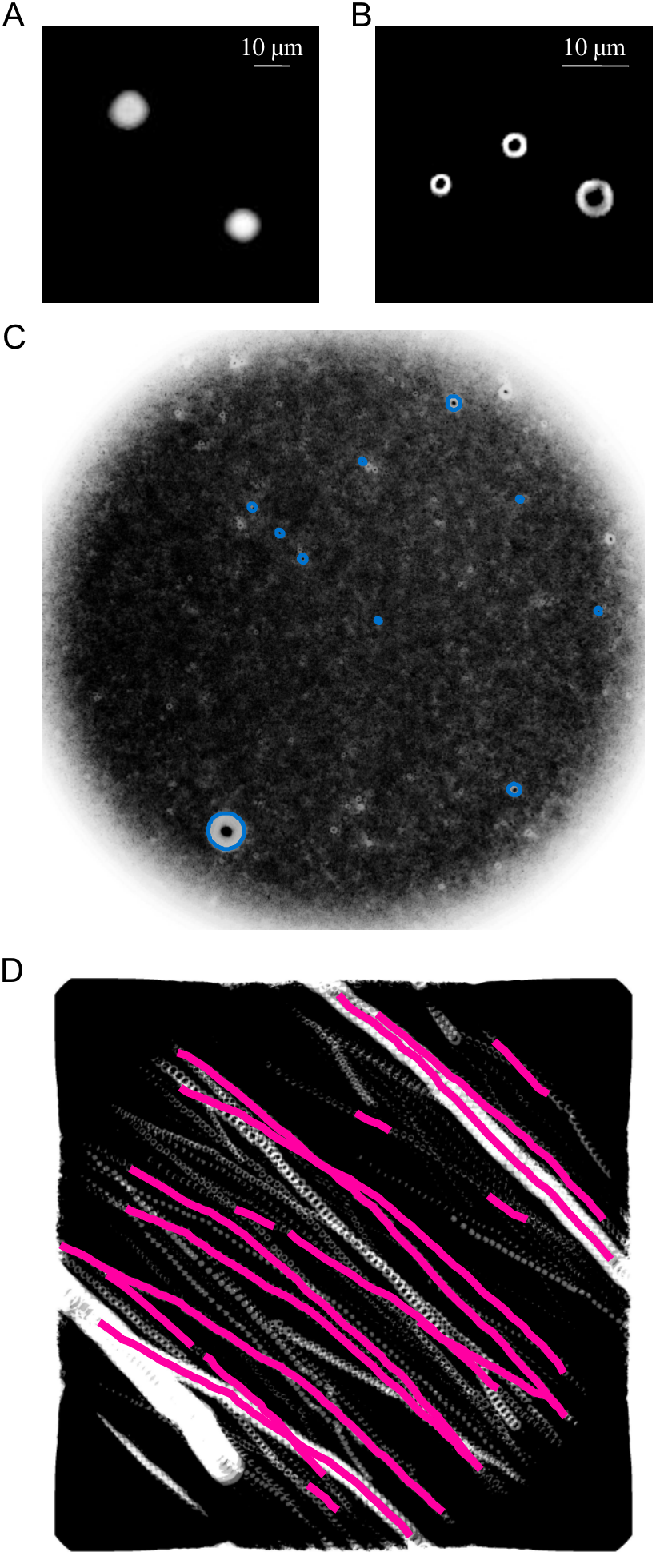}
	\caption{Exemplary image of a bone marrow cells (A) and microbubbles (B); the first frame of a sequence of microscopic frames of microbubbles loaded with SO-Mag5 nanoparticles (gray scale) with the reconstructed objects encircled (blue) (C); an overlay of the series of filtered frames (D) with the reconstructed trajectories (pink)}
	\label{fig:microbubble_tracking}
\end{figure}%

	\item The object positions are subsequently traced throughout the frame sequence by a Dijkstra graph algorithm \citep{Heineman}. The algorithm searches for the best path to any connected object from a single source object. In our case, the edges of the graph algorithm are limited to forward direction and neighboring frames, while additionally, the distance an object can cover between two frames is limited. However, a complete path can span the full image sequence.\\ %Figure \ref{fig:graph_algo} shows a schematic description of the utilized structure. Panels (a) to (c) denote individual frames, while panel (d) shows an overlay of those three frames. All edges between objects of neighboring frames are assigned a weight value which can be based on different quality ratings. For the schematic, we simply chose the distance as a weight parameter. The edges between the objects are plotted as arrows with their weight denoted above them. Starting from one individual object in figure \ref{fig:graph_algo}A, all objects in the next frame are assigned a value identical to the weight of the corresponding edge between the starting object and itself (figure \ref{fig:graph_algo}B). The edge with the smallest weight value is indicated by an orange arrow. This edge connects the start object to the closest object in the next frame. Continuing from this closest object, we repeat the previous step for all objects in the third frame (figure \ref{fig:graph_algo}C). However, we assign a distance value corresponding to the sum of the distances between all three objects. This step is repeated for all other objects of the second frame. The resulting distance value is updated if it is smaller than the previous one. Thereby, instead of the path indicated by the orange arrows, the best path is described by the blue arrows, since the total of their weight is smaller.\\
	In our algorithm, the costs of the edges are comprised of the absolute distance, the relative radius of two objects and the angle of the path as denoted in equations \ref{equ:weights1} to \ref{equ:weights5}. For this purpose, all three properties are calculated beforehand as described above. The ratio $G_{s,r,\varphi}$ of the three cost functions can be chosen freely and the optimal values are highly dependent on the object size, covered distance and distance between the objects and neighboring paths. Typical parameters can be found in table \ref{tab:cell_params1}. The path finding is performed for every object of the first frame and all objects of the subsequent frames which are not already included in a previously found path. Figure \ref{fig:algo_steps}I shows the superimposed frame data and corresponding objects, additionally to the reconstructed trajectory.

	\item After finding all available best paths, the corresponding trajectories are checked for plausibility by comparing the properties of the objects belonging to a trajectory. This includes, on the one hand, comparing the radius and distance between all objects, and, on the other hand, comparing the angles of the segments of each trajectory. Typical values for maximum deviations can be found in table \ref{tab:cell_params3}. If only a few objects of the trajectory do not fit the specifications, the trajectory is split at the corresponding point into several individual trajectories. Figure \ref{fig:microbubble_tracking}D exemplarily shows an overlay of a series of filtered microscopy images and a number of trajectories found by our algorithm.
	
	\item The remaining plausible trajectories, including the split trajectories, can then be evaluated concerning covered distance, velocity and size. From the positions throughout the frame sequence, the recording frame rate and the scale, the velocity and the size of the objects can be deduced. Thereby, we can draw conclusions about certain characteristic properties, e.g. the magnetic moment of the objects. A simplified version of the according equation of motion is given by
	\begin{align}
		\mu\cdot C	&= r\,v\,,	\label{equ:eom_cells}
	\end{align}
	where $\mu$ is the sought after property, $r$ is the object radius and $v$ is the object velocity. $C$ denotes a accumulation of constant terms. In order to determine $\mu$ through the measured data, we use a $\chi^2$-fitting procedure \citep{Press}. For this purpose, we combine the data points $R_{t,m}$ and $s_{t,m}$ to their product
	\begin{align}
		y_{t,m} = R_{t,m}\cdot s_{t,m}\,,
	\end{align}
	where $R_{t,m}$ denotes the radius of the object belonging to one trajectory $t$ and the frame $B_m$ and $s_{t,m}$ is the distance covered by the object between frames $B_m$ and $B_{m+1}$. Using equation \ref{equ:eom_cells}, the model function $M$ and $\chi^2$ read
	\begin{align}
		M		&= \mu \cdot C\,,	\label{equ:model_cells1}\\
		\chi^2	&= \sum_m \left(\frac{y_{t,m} - \mu_t\cdot C}{\sigma_{t,m}}\right)^2\,	\label{equ:model_cells2}
	\end{align}
	with $\mu$ as the free parameter. Minimizing expression \ref{equ:model_cells2} results in an error weighted mean value for the magnetic moment $\mu_t$ of the objects of one trajectory
	\begin{align}
		\mu_t &= \dfrac{1}{c} \frac{\sum\limits_m \dfrac{y_{t,m}}{\left(\sigma^y_{t,m}\right)^2}}{\sum\limits_m \dfrac{1}{\left(\sigma^y_{t,m}\right)^2}}\,.
	\end{align}
	For the estimation of the error $\sigma^y$ on the observable $y$, we assume an error $\sigma^{s,R}$ of 1~px for the distance and object radius. Further assuming a complete correlation between both errors, we gain \citep{Dagostini}
	\begin{align}
		\frac{1}{\left(\sigma_t^\mu\rule{0pt}{1em}\right)^2}	&= c^2 \sum\limits_m \frac{1}{\left(\sigma_{t,m}^y\rule{0pt}{1em}\right)^2}\,,\\
		\left(\sigma_{t,m}^y\rule{0pt}{1em}\right)^2	&= \left(\sigma^{s,R}\rule{0pt}{1em}\right)^2 \left(s_{t,m} + R_{t,m}\right)^2\,.
	\end{align}
\end{enumerate}

\section{Performance and Comparison}

\begin{figure*}
	\centering%
	\includegraphics[width=\textwidth]{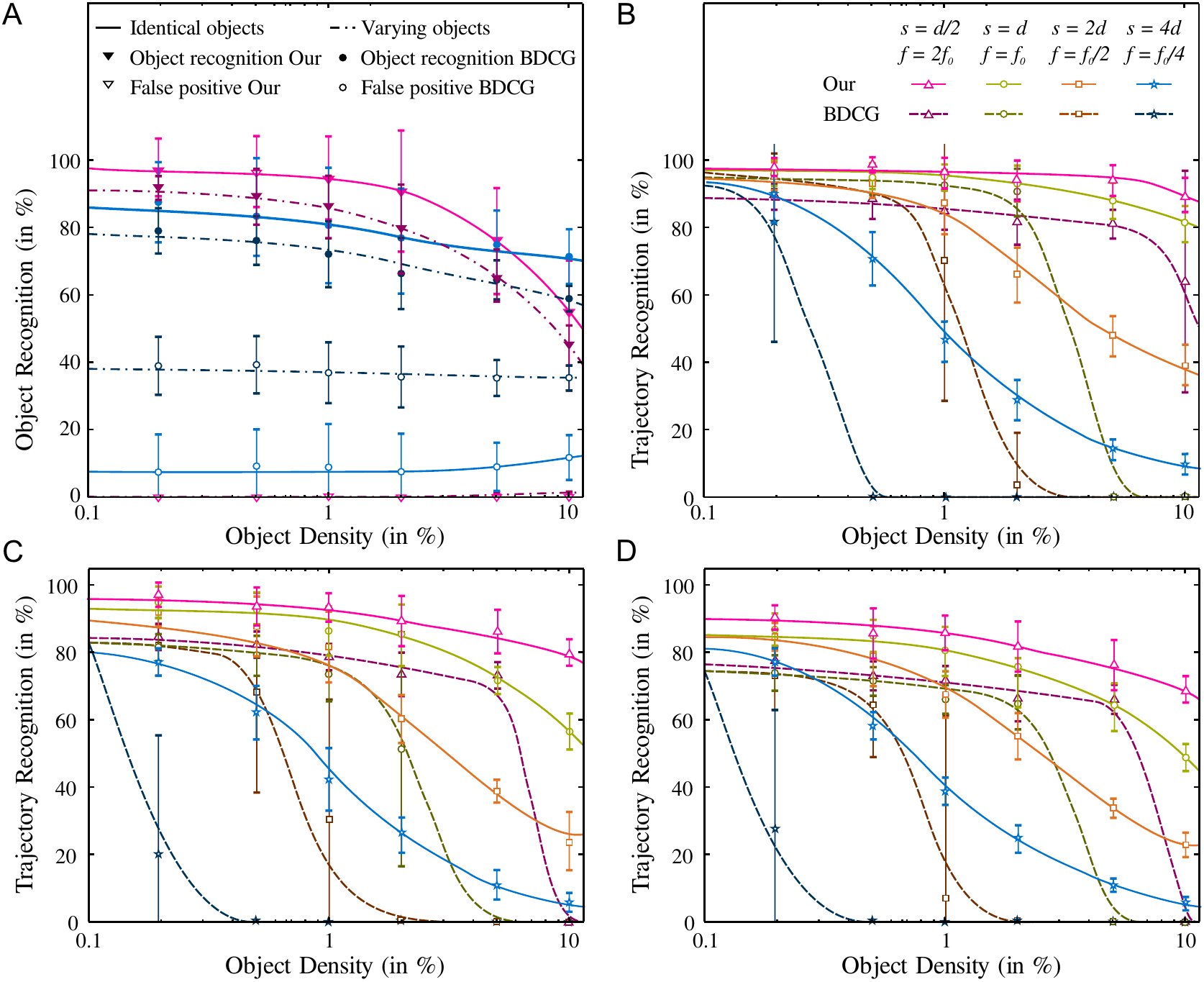}
	\caption{Performance of the cell tracking algorithm in dependence of object density: ratio of recognized objects (A) and ratio of trajectory recognition relative to ratio of object recognition for identical properties (B), properties varying between trajectories (D), properties varying inside trajectory (F); legend for B applies to C and D as well}
	\label{fig:performance}
\end{figure*}%
To verify the results of our tracking algorithm, we performed systematic tests of the recognition rate of the objects and trajectories for different sets of input images. To this purpose, we generated several sets of frames with different object and trajectory properties as well as different object densities.\\
The object density describes the total area of a frame covered by objects and ranges from 0.1 to 10~\% with 20 frames per sequence. The generated objects were allowed to touch, but not to overlap. The mean object diameter $\bar{d}$ was always set to 5~px. The size of the image as well as the number of objects per frame were varied between frame sizes of $125\times125$ to $400\times400$~px and 16 to 80 objects, respectively, to achieve the desired object density. The recognized objects were compared to the previously created objects by position and subsequently, the trajectory composed of those objects was investigated. Here, a fully connected trajectory was rated higher than multiple fragmented pieces of a trajectory.\\
We varied several object and trajectory properties. While the mean object diameter $\bar{d}$ was kept constant, the distance $\bar{s}$ between objects of a trajectory was varied in multiples of the object radius. This is equivalent to changing the frame rate of the recording. For this purpose, we defined the frame rate corresponding to the covered distance \mbox{$\bar{s} = \bar{d} = 5$~px} between frames as $f_0$. Additionally, we varied the deviation of the object properties for objects belonging to the same trajectory and between objects of different trajectories as well as the angle of the trajectory. The standard deviation for those variations was set to 20~\% of the mean value.\\

Figure \ref{fig:performance} shows the results for object and trajectory recognition of our algorithm as well as of the Matlab adaption of \citeauthor{Blair:online} of the tracking algorithm by \citeauthor{Crocker:1996} \citep{Blair:online,Crocker:1996}, in the following called BDCG. Other packages available for testing had poor recognition rates or were not able to reconstruct trajectories. The recognition rates in figure \ref{fig:performance} are displayed as mean value and standard deviation for the tracking of 20 different image sequences.\\
Figure \ref{fig:performance}A shows the amount of recognized objects. %Since the object recognition is performed independently of the trajectory reconstruction, this result does not depend on the weights of the graph $G_{s,r,\varphi}$ or the distance of the objects in a trajectory.
As expected, the recognition rate for identical objects is higher than for varying object sizes. While the recognition rate of our algorithm decreases with increasing object density, the object recognition of BDCG remains constant but approximately 10\% below our recognition rate. The main difference between our algorithm and the BDCG algorithm is the feature recognition step. This explains the decrease in correct object recognition for high object densities since the objects tend to overlap more and more, and no longer fulfill the requirements of our object recognition. The BDCG algorithm has no such limits and therefore finds objects even for high object densities. However, with BDCG we gain a high false positive recognition rate for the same reasons.\\
Figures \ref{fig:performance}B,\,C and D show the proportional amount of recognized trajectories relative to the amount of recognized objects. Colors and markers indicate the frame rate, while color and linestyle are used to differentiate between our and the BDCG algorithm. The legend of figure \ref{fig:performance}B applies also to figures \ref{fig:performance}C and D. While figure \ref{fig:performance}B shows the trajectory recognition for identical objects, moving with same speed in the same direction, figure \ref{fig:performance}C shows the result for trajectories of varying objects, with different size, velocity and direction. However, the object properties within individual trajectories remain constant. Finally, in figure \ref{fig:performance}D shows the trajectory recognition for objects whose properties were varied within a trajectory. This might represent objects drifting slightly out of focus and back, or vibrations causing blurring or lateral displacement.\\
It can be seen that an object distance in the range of the object size or smaller always leads to a higher trajectory recognition. As expected, the success rate of the trajectory recognition decreases with increasing object density and decreasing frame rate. In the special case of identical objects, we have a nearly complete recognition for different objects densitites and high frame rates. %moving with same speed in the same direction (figure \ref{fig:performance}B), the weight on the distance $G_s$ even constrains the recognition rate, since the distance is the most unspecific feature in this case. %The recognition rates remain comparably constant over the different distances / frame rates. The distance weight is far more important for trajectories where the objects vary between trajectories, but not inside them. The reconstructions with $G_s = 0$ in figure \ref{fig:performance}D show less success than other weight combinations. Additionally, the recognition rate is generally lower and decreases faster than the recognition rates in figure \ref{fig:performance}C. 
If the individual objects of a trajectory vary, the recognition rate decreases furthers. The BDCG performs well for high frame rates, but starts to breaks off the computation for lower frame rates, high object densities or varying properties.

\section{Discussion and Conclusion}

As already mentioned, there are many particle and cell tracking algorithms freely or commercially available, but many of those are not suited for our purpose. We presented an algorithm which reliably tracks micrometer-sized objects with deterministic shape and behavior. To this purpose, we used a set of basic algorithms and combined them in a novel way. Though implemented in Matlab, the algorithm does not use any internal frame processing tools or higher functions provided by Matlab, with the exception of the image import. Therefore, it was designed as a platform independent algorithm which can be easily ported to another programing language. Furthermore, the usage of custom and adapted tools for frame filtering, particle recognition and evaluation as well as trajectory recognition is accompanied by an acceleration of the algorithm compared to the usage of generic internal Matlab functions.\\
Though template matching techniques are often used for object recognition, we do not apply those, since we found them to be computationally less efficient for large images and differing object sizes. Also, we do not use shape specific algorithms, e.g. circular Hough transform, but used a generic blob extraction algorithm. Thereby, we separated detection and shape identification and ensured a higher adaptability. Thereby, we can expand the scope of recognized shapes, e.g. to ellipsoids or rectangles, by including additional features in step 3.\\

Our application is the tracking of fast moving objects over large distances along unidirectional trajectories. Therefore, we limit our tracking procedures to gray scale images and are only interested in linearly progressing trajectories. Only movements covering many times the particle radius in distance over relatively short time frames allow for insights into the magnetic behavior. 
In contrast to the previously mentioned cell tracking applications, our algorithm is designed to track homogeneous large distance movement. However, it would also be possible to track movement following different behavior as long as we are able to describe it sufficiently by simple mathematical expressions. The weights for the individual cost functions of the graph are given by the user and can thereby be varied or disabled as necessary for the recorded frame set, e.g. disabling the angular weight in step 5 would enable us to track Brownian movement similar to other tracking algorithms. Additionally, the reconstruction step of the algorithm can be easily expanded by including additional cost functions in step 4, e.g. shape or orientation. Thereby, we achieve a variability in the assessment of the movement (direction) and the object similarities.\\
Since many cell types exhibit a nearly monodisperse size distribution, the identification solely via the object size and absolute distance is not adequate for our purpose. Therefore we also consider the relative angle between the individual objects. Since the objects move unidirectional towards the magnetic field source, the angle between all objects of a trajectory has to be comparable. Furthermore, we do not only compare the values of size, distance and angle for subsequent objects, but also compare those values over the whole trajectory to exclude a merging or switching of adjacent trajectories.\\
We do not consider object interactions or dis- and reappearances of objects. If an object is temporarily out of focus or overlaps with another, we simply ignore those steps and start a new trajectory after the incident. This works well as long as disappearances do not happen very often and the trajectories are thereby, not broken into too short fragments. In a similar way, we split trajectories at irregular points like significant changes in size or angle. However, having multiple short trajectories instead of one long trajectory can impair the statistics of the result. Assuming that breaks in the trajectory occur mostly due to changes in the trajectory itself, e.g. due to external influences like vibrations, those disturbances are applied to the whole frame. This interrupts all current trajectories in the same way and the fragments contribute equally to the statistic. Thereby, this influence is acceptable.\\

As shown above, our algorithm performs very well and surpasses the BDCG algorithm for object and especially for trajectory recognition. The feature based object recognition is very useful for real microscopy images, since it discards all non-matching objects. In our case, those were mostly objects which were not completely in focus or adhering cells or microbubbles. Those complexes of two or more objects would, on the one hand, not follow the same velocity profile and, on the other hand, falsify results by including much larger object sizes.\\

\section{Acknowledgements}
This work is supported by the ``Deut\-sche Forschungsgesellschaft'' (DFG) within the Research Unit 917 ``Nanoparticle-based targeting of gene- and cell-based therapies''. Microbubbles were kindly provided by H. Mannell (Walter Brendel Centre of Experimental Research, Ludwig-Maximilians-Universit\"at M\"unchen). Bone marrow cells were provided by W. R\"oell and prepared by A. Ottersbach (Klinik f\"ur Herzchirurgie and Institut f\"ur Physiologie 1, Universit\"at Bonn). The author would like to thank T. H. for his helpful comments and discussion.

%\bibliography{Paper}   % name your BibTeX data base

\end{document}